\documentclass{article}

\usepackage[preprint,nonatbib]{neurips_2021}
\usepackage{graphicx}

\usepackage[utf8]{inputenc} 
\usepackage[T1]{fontenc}    
\usepackage{url}            
\usepackage{booktabs}       
\usepackage{amsfonts}       
\usepackage{nicefrac}       
\usepackage{microtype}      
\usepackage{xcolor}         
\usepackage{todonotes}      
\usepackage[ruled,vlined]{algorithm2e}
\usepackage{listings}
\usepackage{wrapfig}
\usepackage{hyperref}       
\definecolor{mygreen}{rgb}{0,0.6,0}
\newsavebox{\codebox}

\title{Temperature as Uncertainty in Contrastive Learning}

\begin{document}

\author{
    Oliver Zhang$^{1}$, Mike Wu$^{1}$, Jasmine Bayrooti$^{1}$, Noah Goodman$^{1,2}$ \\
    Department of Computer Science$^{1}$ and Psychology$^{2}$\\
    Stanford University\\
    Stanford, CA 94303 \\
    \texttt{\{ozhang, wumike, jbayrooti, ngoodman\}@stanford.edu}
}

\maketitle

\begin{abstract}

Contrastive learning has demonstrated great capability to learn representations without annotations, even outperforming supervised baselines. However, it still lacks important properties useful for real-world application, one of which is uncertainty. 
In this paper, we propose a simple way to generate uncertainty scores for many contrastive methods by re-purposing temperature, a mysterious hyperparameter used for scaling.
By observing that temperature controls how sensitive the objective is to specific embedding locations, we aim to learn temperature as an input-dependent variable, treating it as a measure of embedding confidence. We call this approach ``Temperature as Uncertainty'', or TaU. 
Through experiments, we demonstrate that TaU is useful for out-of-distribution detection, while remaining competitive with benchmarks on linear evaluation. 
Moreover, we show that TaU can be learned on top of pretrained models, enabling uncertainty scores to be generated post-hoc with popular off-the-shelf models.
In summary, TaU is a simple yet versatile method for generating uncertainties for contrastive learning. Open source code can be found at: 
\url{https://github.com/mhw32/temperature-as-uncertainty-public}.

\end{abstract}

\section{Introduction}
\label{sec:introduction}


\begin{wrapfigure}{r}{0.6\linewidth}
    \centering
    \includegraphics[width=0.45\linewidth]{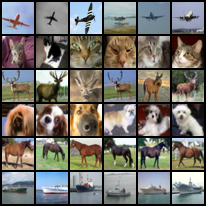}
    \includegraphics[width=0.45\linewidth]{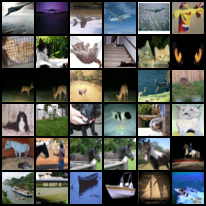}
\caption{CIFAR10 Images on the left have high TaU certainty while images on the right have low TaU certainty.}
\label{fig:pitch}
\end{wrapfigure}

Representation learning through contrastive objectives has recently broken new ground, matching the performance of fully supervised methods on image classification \cite{hjelm2018learning,he2020momentum,misra2020self,chen2020simple,chen2020improved,grill2020bootstrap,chen2021exploring,zbontar2021barlow}. 
While contrastive learning has shown strong practical results, it still lacks some important properties useful for real-world decision-making. One such property is uncertainty, which plays an important role in intelligent systems recognizing and preventing errors. For example, uncertainty can be leveraged to find anomalies that are out-of-distribution (OOD), on which a model's predictions may be degraded or entirely out-of-place.
However, current contrastive frameworks do not provide any indication of uncertainty as they learn one-to-one mappings from inputs to embeddings.

Our work uses the temperature parameter to estimate the uncertainty of an input. While almost all contrastive frameworks include  temperature in the objective, it historically has remained relatively unexplored compared to work on negative samples \cite{wu2020conditional,xie2020delving}, stop gradients \cite{grill2020bootstrap,chen2021exploring,zbontar2021barlow}, and transformation families \cite{tamkin2020viewmaker,tian2020makes}.
Recently, it has been shown that smaller temperature increases the model's penalty on difficult negative examples \cite{wang2021understanding}. With this intuition, we make temperature a learned, input-dependent variable. High temperature is tantamount to the model declaring that a training input is difficult. Temperature, therefore, can be viewed a form of uncertainty.
We call this simple extension to the contrastive objective, ``Temperature as Uncertainty'' or TaU for short.

On benchmark image datasets, we show that TaU is useful for out-of-distribution detection, outperforming baseline methods for extracting uncertainty such as ensembling or Bayesian posteriors over weights. We also show that one can easily derive uncertainty on top of pretrained representations, making this approach widely applicable to existing model checkpoints and infrastructure.

\section{Temperature as Uncertainty}
\label{sec:model}

To start, we give a brief overview of contrastive learning to motivate the approach. Suppose we have a dataset $\mathcal{D} = \{x_i\}_{i=1}^n$ of i.i.d image samples from $p(x)$, a distribution over a space of natural images $X$.
Let $\mathcal{T}$ be some family of image transformations, $t: X \rightarrow X$, equipped with a distribution $p(t)$.
The common family of visual transformations includes a random mix of cropping, color jitter, gaussian blurring, and horizontal flipping \cite{wu2018unsupervised,tian2020contrastive,zhuang2019local,bachman2019learning,he2020momentum,chen2020simple}.

Define an encoding function $f: X \rightarrow \mathbf{S}^{d-1}$ that maps an image to a L$_2$-normalized representation. Let $f$ be parameterized by a deep neural network. The contrastive objective for the $i$-th example is:
\begin{equation}
\mathcal{L}(x_i) = \mathbf{E}_{t,t',t_{1:k} \sim p(t)} \mathbf{E}_{x_{1:k} \sim p(x)} \left[ \log \frac{e^{f(t(x_i)) \cdot f(t'(x_i))/\tau}}{\frac{1}{k}\sum_{j\in\{1:k\}} e^{f(t(x_i)) \cdot f(t_j(x_j))/\tau}} \right]
\label{eq:contrastive}
\end{equation}
where $x_{1:k} = \{ x_1, \ldots, x_k \}$ represents $k$ i.i.d. samples, and $\tau$ is the temperature. We call transformations of the same image ``positive examples'' and transformations of different images ``negative examples''.
We chose to present Eq.~\ref{eq:contrastive} in a very general form based on the noise contrastive \cite{oord2018representation,gutmann2010noise} lower bound to mutual information  \cite{hjelm2018learning,poole2019variational,wu2020mutual,tian2020makes}, although many popular frameworks like SimCLR \cite{chen2020simple} and MoCo-v2 \cite{chen2020improved} can be directly derived from Eq.~\ref{eq:contrastive}.

\begin{lrbox}{\codebox}
\begin{lstlisting}[
  language=Python,
  basicstyle=\tiny,
  commentstyle=\color{mygreen},
]
# g: TaU encoder networks
# x: minibatch of images
# scale: constant (e.g. 0.1)

# encode augmentation and return
# embedding and temperature
# inv_tau is used for stability
emb1,tau1=g(aug(x))
emb2,tau2=g(aug(x))
emb1=norm(emb1)  # L2 norm
emb2=norm(emb2)
emb=cat(emb1,emb2)
# rescale for numerical stability
tau1 = sigmoid(tau1) / scale
tau2 = sigmoid(tau2) / scale
tau=cat(tau1,tau2)
# compute similarities
pos_dps = sum(emb1*emb2) * tau1
neg_dps = emb@emb.T * tau
# mask out identity comparisons
neg_dps = simclr_mask(neg_dps)
# parametric log softmax
loss=-pos_dps+logsumexp(neg_dps)
\end{lstlisting}
\end{lrbox}
\begin{wrapfigure}{L}{0.34\textwidth}
  \begin{minipage}{0.34\textwidth}
    \begin{algorithm}[H]
      \usebox{\codebox}
      \caption{TaU + SimCLR}
      \label{fig:algos}
    \end{algorithm}
  \end{minipage}
\end{wrapfigure}

Since Eq.~\ref{eq:contrastive} uses a dot product as a distance function, the role of temperature, $\tau$ is to scale the sensitivity of the loss function \cite{wang2021understanding}. A $\tau$ closer to 0 would accentuate when representations are different, resulting in larger gradients. In the same vein, a larger $\tau$ would be more forgiving of such differences. In practice, varying $\tau$ has a dramatic impact on embedding quality. Traditionally, there are fixed values for $\tau$ that the authors of a contrastive framework have painstakingly tuned.


We decide to learn an input-dependent temperature. In accordance with previous observations, learning an input-dependent temperature would amount to an embedding sensitivity for every input. In other words, \textit{a measure of representation uncertainty}. Inputs with high temperature suggest more uncertainty as the objective is more invariant to displacements, whereas inputs with low temperature suggest less uncertainty as the objective is more sensitive to changes in embedding location. 

Implementing this idea is very straightforward. We can replace $\tau$ in Eq.~\ref{eq:contrastive} with $\tau(t(x_i))$, overloading notation to define a mapping $\tau: X \rightarrow (M, \infty)$ for some lower bound $M$. We call this new objective TaU, or \textbf{T}emperature \textbf{a}s \textbf{U}ncertainty. In practice, we edit the encoder network $f(x)$ to return $d+1$ entries, the first $d$ of which are the embedding of $x$, and the last entry being the uncertainty for $x$. A sigmoid is used to bound $\tau$ to be positive and a fixed constant is used to set a lower bound $M$ for stability. See Algo.~\ref{fig:algos} for pseudo-code.

\section{Related Work}
\label{sec:related}
\textbf{Learned Temperature}$\quad$ Methods which learn temperature can be found in supervised learning \cite{zhang2018heated,agarwala2020temperature}, model calibration \cite{guo2017calibration,neumann2018relaxed}, language supervision \cite{radford2021learning}, and few-shot learning \cite{oreshkin2018tadam,rajasegaran2020self}. In most of these approaches, temperature is treated as a global parameter when it is learned, not as a function of the input as in TaU. 
The only example, to the best of our knowledge, of learned temperature as a function of the input is \cite{neumann2018relaxed}, which uses temperature for calibration. Instead, we use temperature in the context of self-supervised learning and apply it to OOD detection. 

\textbf{Uncertainty in Deep Learning}$\quad$ There is a rich body of work in adding uncertainty to deep learning models \cite{gawlikowski2021survey,abdar2021review}, of which we highlight a few. Most straightforward is ensembling of neural networks \cite{tao2019deep}, where multiple copies are trained with different parameter initializations to find many local minima. Further work attempts to enforce ensemble diversity for more variance \cite{liu2019deep,abbasi2020toward}. Another popular approach of uncertainty is through Bayesian neural networks \cite{goan2020bayesian}, of which the most practical formulation is Monte Carlo dropout \cite{gal2016dropout}. This approach frames using dropout layers during training and test time as equivalent to sampling weights from a posterior distribution over model parameters. Finally, most relevant is ``hedged instance embeddings'' \cite{oh2018modeling}, which edits the contrastive encoder $f$ to map an image to a Gaussian distribution, rather than a point embedding. The primary drawbacks of this approach are (1) computational cost as it requires multiple samples, and (2) it is not proven to work in high dimensions. In our experiments, we compare these baselines to TaU.

\textbf{OOD Detection}$\quad$ Existing OOD algorithms mostly derive outlier scores on top of predictions made by large supervised neural networks trained on the inlier dataset, such as using the maximum softmax probability \cite{hendrycks2016baseline}, sensitivity to parameter perturbations \cite{liang2017enhancing}, or Gram matrices on activation maps \cite{sastry2019detecting} as the outlier score.
While these methods work very well, reaching near ceiling performance, they require human annotations, which may not be available. 


\section{Experiments}
\label{sec:experiments}

A primary application of uncertainty is to find abnormal or anomalous inputs. We aim to show that using TaU temperatures as uncertainty is effective for out-of-distribution (OOD) detection \cite{hendrycks2016baseline, liang2017enhancing, sastry2019detecting}, with the added bonus of sacrificing little to no performance on downstream tasks.

\textbf{OOD Detection}$\quad$
\label{sub:ood}
We study OOD detection, where inputs from an anomalous distribution are fed to a trained model. A well-performing metric should assign high uncertainty to these OOD inputs, thereby making it possible to classify whether an input is OOD.

We train TaU on CIFAR10 as the inlier dataset, and consider three different OOD sets: CIFAR100, SVHN, and TinyImageNet. We note that CIFAR10 and CIFAR100 are very similar in distribution, whereas SVHN is the most dissimilar. To measure performance, we compute AUROC on correctly classifying an example as OOD or not. We compare TaU to several baselines. Assuming unrestricted computational power and memory, one expensive procedure to derive uncertainties is to fit a k-nearest neighbors algorithm on the entire training corpus, and treat the average distance of a new example to $k$ neighbors as an uncertainty score. We try out $k=[1, 3, 10, 32, 100]$ with $k=10$ working the best. This serves as an upper bound on performance. For other baselines, please refer to Sec.~\ref{sec:related}.

\begin{table}[h]
    \centering
    \small
    \caption{\textbf{Downstream Out-of-Distribution Detection}: comparison of TaU to several popular baselines for uncertainty on deep neural networks. Out-of-distribution AUROC is reported.}
    \begin{tabular}{lccc}
        \toprule
        \textbf{Method} (CIFAR10) & \textbf{CIFAR100} & \textbf{SVHN} & \textbf{TinyImageNet} \\
        \midrule
        TaU + SimCLR & $\textbf{0.746}$ & $0.964$ & $\textbf{0.760}$ \\
        TaU + MoCo-v2 & $0.728$ & $\textbf{0.968}$ & $0.746$ \\
        SimCLR + kNN & $\textbf{0.746}$ & $0.829$ & $0.756$ \\
        MoCo-v2 + kNN & $0.712$ & $0.800$ & $0.726$ \\
        SimCLR + MC Dropout \cite{gal2016dropout} & $0.504$ & $0.684$ & $0.512$ \\
        Supervised + MC Dropout \cite{gal2016dropout} & $0.659$ & $0.745$ & $0.722$ \\
        Hedged Instance Embedding \cite{oh2018modeling} & $0.509$ & $0.834$ & $0.508$ \\
        Ensemble of 5 SimCLRs & $0.532$ & $0.525$ & $0.513$ \\
        \bottomrule
    \end{tabular}
    \label{tab:ood}
\end{table}

From Table~\ref{tab:ood}, we observe that on CIFAR100 and TinyImageNet -- two image corpora with similar content as CIFAR10 -- TaU outperforms (or matches) all baselines, though only surpassing SimCLR + kNN by a small margin of 0-2\%.  However, for SVHN -- an image corpus very different in content to CIFAR10 -- TaU outperforms all baselines by at least 13\%. In fact, we find most baselines do not generalize well to contrastive learning, as many perform near chance AUROC. Even prior methods specifically for contrastive uncertainty \cite{oh2018modeling} do not consistently perform well. The exception is using kNN \textit{with the caveat} that the OOD set was not too far from the training set. Nearest neighbors fundamentally relies on a good distance function, which is achievable when the OOD input can be properly embedded. But in cases when we are truly OOD, it may not be clear where to embed an anomalous image. In these cases, as with SVHN, kNN approaches struggle.

\textbf{Linear Evaluation}$\quad$
Although we have shown that TaU uncertainties can detect OOD inputs, they would not be much good if it came at a large cost of performance on downstream tasks. 
To show this is not the case, we measure performance through both linear evaluation \cite{chen2020simple} and  k-nearest neighbors on the training set \cite{zhuang2019local} (where the predicted label for a test example is the label of the closest example in the training set). Please refer to the appendix for experiment details.

\begin{table}[h]
    \centering
    \small
    \caption{\textbf{Downstream Image Classification}: mean and standard deviation in accuracy are measured over three runs with different random seeds. The best performing models are bolded.}
    \begin{tabular}{lcc}
        \toprule
        \textbf{Method} & \textbf{kNN Eval} & \textbf{Linear Eval} \\
        \midrule
        TaU + SimCLR & $0.762 \pm 0.001$ & $0.750 \pm 0.003$ \\
        TaU + MoCo-v2 & $0.709 \pm 0.004$ & $0.690 \pm 0.004$ \\
        SimCLR & $\mathbf{0.787} \pm 0.004$  & $\mathbf{0.775} \pm 0.002$ \\
        MoCo-v2 & $0.734 \pm 0.004$ & $0.720 \pm 0.005$ \\
        \bottomrule
    \end{tabular}
    \label{tab:acc}
\end{table}

From Table~\ref{tab:acc}, we observe TaU to perform only slightly worse than their deterministic counterparts, with a small reduction of 2-3 percentage points on both linear and k-nearest neighbors evaluation. While there is a non-zero cost to adding uncertainty, we believe that trading a few percentage points for a measure of confidence is practically worthwhile.


\textbf{Uncertainty on Pretrained Models}$\quad$
We next show that TaU can be finetuned on top of pretrained models, enabling uncertainties to be generated post-hoc on popular off-the-shelf checkpoints. Specifically, we finetune on supervised, SimCLR, BYOL, and CLIP embeddings. All models were pretrained using ResNet50 on ImageNet with the exception of CLIP, which uses a ViT \cite{dosovitskiy2020image}. We finetune TaU uncertainties for 40 epochs, and all images were reshaped to 224 by 224 pixels.

\begin{table}[h]
    \centering
    \small
    \caption{\textbf{Out-of-Distribution Detection using Pretrained Embeddings}: using TaU to generate uncertainties for several pretrained models. Out-of-distribution AUROC is reported.}
    \begin{tabular}{lccccccc}
        \toprule
        \textbf{Method} (ImageNet) & \textbf{CIFAR10} & \textbf{CIFAR100} & \textbf{SVHN} & \textbf{TinyImgNet} & \textbf{LSUN} & \textbf{COCO} & \textbf{CelebA} \\
        \midrule
        TaU + Supervised & \textbf{0.913} & \textbf{0.874} & \textbf{0.978} & \textbf{0.771} & 0.657 & 0.458 & 0.657 \\
        TaU + SimCLR \cite{chen2020simple} & 0.823 & 0.870 & 0.968 & 0.747 & 0.552 & 0.554 & 0.717 \\
        TaU + BYOL \cite{grill2020bootstrap} & 0.763 & 0.808 & 0.955 & 0.686 & 0.471 & 0.497 & 0.840\\
        TaU + CLIP \cite{radford2021learning} & 0.056 & 0.044 & 0.071 & 0.154 & \textbf{0.779} & \textbf{0.579} & \textbf{0.883}\\
        \bottomrule
    \end{tabular}
    \label{tab:pre}
\end{table}

Table~\ref{tab:pre} reports AUROC for OOD detection for a wide survey of outlier datasets. We find that for supervised, SimCLR, and BYOL embeddings, the learned TaU uncertainties are largely able to classify OOD inputs. The exception is with COCO, likely due to a close similarity with ImageNet data points. However, CLIP surprisingly faces the opposite problem with low OOD scores for most datasets but outperforming in COCO and LSUN. Further work could explore whether CLIP's behavior is due to differences in objective, architecture, or training. 

\section{Limitations and Future Work}
\label{sec:limitations}

We presented TaU, a simple method for adding uncertainty into contrastive learning objectives by repurposing temperature as uncertainty. In our experiments, we compared TaU to existing benchmark algorithms and found competitive downstream performance, in addition to TaU uncertainties being useful for out-of-distribution detection. We then demonstrated how uncertainty can be added to already trained model checkpoints, enabling practitioners to reuse computation. 

We discuss an important limitation: our approach is restricted to contrastive algorithms built on NCE. Other approaches, such as SimSiam \cite{chen2021exploring}, BYOL \cite{grill2020bootstrap}, and Barlow Twins \cite{zbontar2021barlow}, replace negative examples entirely with stop gradients, where we find limited success with TaU. Future work can also explore TaU-like techniques for detecting corrupted or adversarial examples.


\bibliographystyle{plain}
\bibliography{reference}

\newpage
\appendix

\section{Training Hyperparameters}

\paragraph{Pretraining} 

For all models, we use a representation dimensionality of 128. We use the LARS  optimizer \cite{you2017large} with learning rate 1e-4, weight decay 1e-6, batch size 128 for 200 epochs, as described in \cite{chen2020simple}. For baseline models (no uncertainty), we use a fixed temperature $\tau = 0.1$. For MoCO-V2, we use $K=65536$ negative samples with a momentum of 0.999 for updating the memory queue. We use the same optimizer as described above but with learning rate 1e-3. For CIFAR10, all images are resized to 32x32 pixels; For ImageNet, all images are resized to 256x256 pixels (with 224x224 cropping size). During pretraining, we use random resized crop, color jitter, random grayscale, random gaussian blur, horizontal flipping, and pixel normalization (with ImageNet statistics). During testing, we only center crop and do pixel normalization. For encoders we train from scratch, we use ResNet18 \cite{he2016deep} for encoders $f$. We adapted the Pytorch Lightning Bolts implementations of SimCLR and MoCo-v2, found here: \url{https://github.com/PyTorchLightning/lightning-bolts}.

For the larger models, we downloaded existing SimCLR (ResNet50) checkpoints trained on ImageNet from \url{https://github.com/google-research/simclr} and converted it to PyTorch checkpoints using \url{https://github.com/Separius/SimCLRv2-Pytorch}. 

\paragraph{Downstream Classification}

We freeze encoder parameters, remove the final L$_2$ normalization, and append a 2-layer MLP with hidden dimension of 128 and ReLU nonlinearity. For optimization, we use SGD with batch size 256, learning rate 1e-4, weight decay 1e-6, and cosine learning rate schedule that drops at epoch 60 and 80, with a total of 100 epochs.

\paragraph{Optimization Stability} 

When optimizing the TaU objective, we found optimization instability where $\tau(z)$ would either collapse to 0 or converge to $\pm\inf$ if left unbounded. We found it crucial to employ some tricks for optimization stability. First, we follow Neumann et al. and have our network predict some $\alpha(z) = \frac{1}{\tau(z)}$ instead of $\tau(z)$ directly \cite{neumann2018relaxed}. This changes the training dynamics but does not change the underlying equation. Second, we bound $\alpha(z)$ to between 0 and 1 using a sigmoid function. Finally, we divide $\alpha(z)$ by $0.1$, which helps initialize the temperature to be in the same range as fixed-temperature models.
When using the uncertainty for out-of-distribution detection, we found that using the pre-sigmoid $\alpha(z)$ worked much better than the post-sigmoid $\alpha(z)$, as the differences between post-sigmoid values became indistinguishable using float32.

\end{document}